\pdfoutput=1
\documentclass[11pt]{article}

\usepackage[]{acl}

\usepackage{times}
\usepackage{latexsym}
\usepackage[T1]{fontenc}
\usepackage[utf8]{inputenc}

\usepackage{microtype}

\usepackage{amssymb}
\usepackage{amsmath}
\usepackage{graphicx}
\usepackage{float}
\usepackage{algorithm}
\usepackage{algpseudocode}
\usepackage{booktabs}
\usepackage{multirow}
\usepackage{svg}
\usepackage{enumitem}
\usepackage{hyperref}

\title{Deterministic Reversible Data Augmentation \\for Neural Machine Translation}

\author{
\textbf{Jiashu Yao}$^{1}$ \quad
\textbf{Heyan Huang}$^{1}$ \quad
\textbf{Zeming Liu}$^{2}$ \quad
\textbf{Yuhang Guo}$^{1}$\thanks{\ \ Corresponding Author}
\\
$^{1}$School of Computer Science and Technology, Beijing Institute of Technology \\
$^{2}$School of Computer Science and Engineering, Beihang University \\
\texttt{\{yaojiashu, hhy63, guoyuhang\}@bit.edu.cn, zmliu@buaa.edu.cn}
}

\begin{document}
\maketitle
\begin{abstract}
Data augmentation is an effective way to diversify corpora in machine translation, but previous methods may introduce semantic inconsistency between original and augmented data because of irreversible operations and random subword sampling procedures. To generate both symbolically diverse and semantically consistent augmentation data, we propose Deterministic Reversible Data Augmentation (DRDA), a simple but effective data augmentation method for neural machine translation. DRDA adopts deterministic segmentations and reversible operations to generate multi-granularity subword representations and pulls them closer together with multi-view techniques. With no extra corpora or model changes required, DRDA outperforms strong baselines on several translation tasks with a clear margin (up to 4.3 BLEU gain over Transformer) and exhibits good robustness in noisy, low-resource, and cross-domain datasets. The relevant code is available at \url{https://github.com/BITHLP/DRDA}.
\end{abstract}

\section{Introduction}
\label{sec:intro}

\begin{figure*}[!htb]
    \centering
    \includegraphics[width=0.95\linewidth]{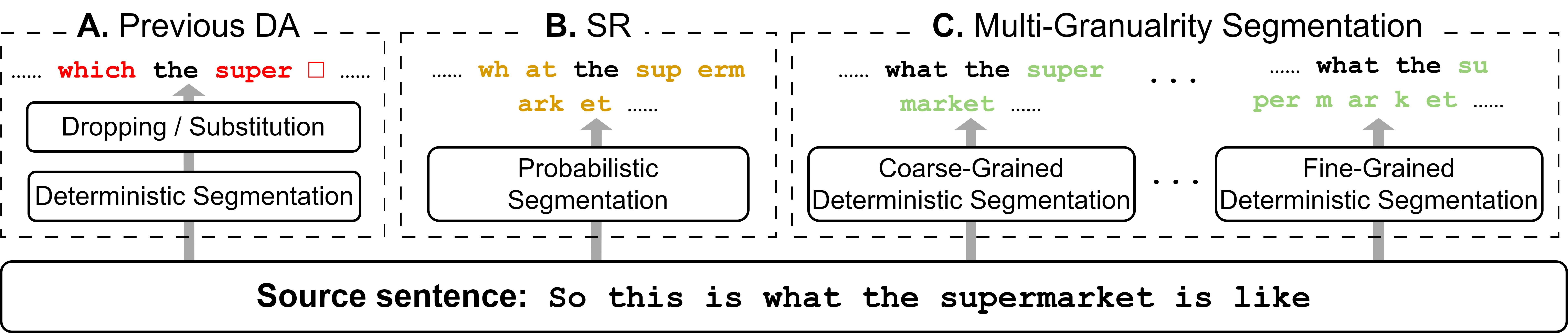}
    \caption{Subword piece sequences generated by previous data augmentation (A), subword regularization (B), and multi-granularity segmentation (C) representing the same source sentence. $\Box$ denotes an empty subword (a zero vector). Previous data augmentation methods result in semantic loss (red texts), subword regularization may sample inappropriate subwords (yellow texts), while multi-granularity segmentation generates symbolically diverse and semantically consistent augmentation data (green texts).}
    \label{fig:tmp}
\end{figure*}

Recent neural machine translation (NMT) models have led to dramatic improvements in translation quality. However, the powerful learning and memorizing ability of these models also leads to poor generalization and vulnerability to small perturbations like misspelling and paraphrasing \cite{belinkov2017synthetic, cheng2020advaug}.

A common solution to perturbation vulnerability is data augmentation \cite{sennrich2016improving,cheng2016semi}, which is to create massive virtual training data with diverse symbolic representations under the premise of ensuring semantic consistency \cite{cheng2019robust,cheng2020advaug}. Symbolic diversity emphasizes that original and augmented data should differ significantly in token sequences, and semantic consistency requires that the two should be semantically similar. Previous data augmentation methods employ irreversible substitutions, like direct dropping or replacing discrete tokens to generate diverse data (Figure \ref{fig:tmp} A). Despite being able to improve data diversity, these augmentation operations are not reversible, and will inevitably introduce semantic loss to original texts, thus compromising the semantic consistency between original and augmented data.

Yet another way to generate diverse augmentation data without employing irreversible operations is subword regularization \cite{kudo2018subword,provilkov2020bpe}. Subword regularization adopts random segmentations to sample subwords probabilistically thus generating diverse data. These methods are reversible because of the inherent reversibility of segmentations. However, due to the random sampling procedure of segmentation, they may adopt inappropriate subword segmentations (e.g., "\texttt{sup erm ark et}" in Figure \ref{fig:tmp} B). These sub-optimal segmentations may result in semantic perturbations and do damage to semantic consistency.

To summarize, previous methods have difficulty in completely retaining the semantics from corruption when diversifying the texts because of irreversible augmentation operations and probabilistic subword sampling.

To generate symbolically diverse and semantically consistent data, we propose Deterministic Reversible Data Augmentation (DRDA), a simple but effective augmentation approach. DRDA augments source sentences with their token representations in different granularities as shown in Figure \ref{fig:tmp} C. These representations are symbolically diverse, but also syntactically correct and semantically complete thanks to the reversible and deterministic segmentations in the multi-granularity segmentation process. To make full use of the semantic identity among all multi-granularity representations of one sentence, we also leverage the multi-view techniques in training to pull these representations closer together.

We conduct extensive experiments of different languages and scales and find that DRDA gains consistent improvements over strong baselines with clear margins. To further understand the factors that make DRDA work, we conduct insightful analyses of the effects DRDA imposed on semantic consistency, subword frequency, and subword semantic composition. We combine the empirical and theoretical verification of the consistency and offer a subword-level explanation of the mechanism of multi-granularity segmentations and multi-view techniques.

Our contributions are summarized as follows:

\begin{itemize}%[noitemsep]
    \item We propose DRDA that exclusively employs deterministic reversible operations to generate diverse augmentation data without introducing semantic noise.
    \item We conduct extensive experiments and verify the high effectiveness of DRDA.
    \item To investigate the factors that make DRDA work, we combine empirical and theoretical analyses and offer insightful explanations.
\end{itemize}

\section{Related Work}

\paragraph{Augmentation methods}

Besides continuous approaches \cite{wei2022learning}, data augmentation can be categorized into back-translation like methods \cite{sennrich2016improving, edunov2018understanding, nguyen2020data} and token substitution methods. DRDA is an instance of the latter category.

Several substitution methods uniformly select a word or token in a sentence and perform deletion or substitution \cite{zhang2020token, shen2020simple, wang2018switchout, norouzi2016reward, gao2022bi}. \citet{cheng2019robust, cheng2020advaug} constrained the substitution of a word in a small subset of synonyms, thus improving the semantics consistency. \citet{kambhatla2022cipherdaug} viewed the original corpus as plain text and applies a rotation encryption as data augmentation. Unlike previous methods, introducing multi-granularity takes advantage of the reversible nature of segmentation and causes no semantic loss.

\paragraph{Subword regularization}

The de-facto subword method, BPE \cite{sennrich2016neural}, still suffers from sub-optimality \cite{bostrom2020byte}. To overcome this sub-optimality, several subword regularization approaches are proposed. \citet{kudo2018subword} and \citet{provilkov2020bpe} presented subword regularization by modelling segmentation ambiguity. \citet{wang2021multi} integrated BPE and BPE-Drop by enforcing the consistency using multi-view subword regularization, \citet{wu2020sequence} and \citet{kambhatla2022auxiliary} combined BPE in \texttt{SentencePiece} and \texttt{subword-nmt} together to obtain regularization effects. DRDA is distinct from all the random sampling segmentation methods, as the augmentation data is generated deterministically. The determinism helps alleviate less reasonable segmentation, while achieving regularization effects as well.

In addition, other researches put efforts into taking advantage of multi-granularity representations, which can also be viewed as a subword regularization. \citet{li2020flat} and \citet{gao2020character} adopted word lattice and convolutions of different kernel sizes respectively, \citet{chen2018combining} and \citet{li2022learning} combined levels of representation scales, \citet{hao2019multi} modified self-attention module to introduce phrase modeling. Unlike these methods, DRDA requires no modification to model architectures and can be applied to universal tasks.

\section{Background: Subword Segmentation}
\label{sec:seg}

Subword segmentation models the probability of token sequence $ \mathbf{x} = x_1, x_2, ..., x_{m}$ given a source sentence $ \mathbf{s} $. Previous deterministic subword segmentations choose the most probable sample:

\begin{equation}
\begin{aligned}
    \mathbf{x}^* &= \mathop{\arg\max}\limits_{\mathbf{x}} P_{seg}(\mathbf{x} | \mathbf{s};p)\\
    &= \mathop{\arg\max}\limits_{\mathbf{x} \in V_p} P_{seg}(\mathbf{x} | \mathbf{s}),
\label{equ:subword-max}
\end{aligned}
\end{equation}
where $p$ is the size of the vocabulary (a set of subword candidates), and each token $x_i$ ($i \in \{1,2,...,m\}$) is selected from vocabulary $V_p$. For example, Byte Pair Encoding (BPE) assigns $P(\mathbf{\hat{x}} | \mathbf{s};p)=1$ when $\hat{x}$ is obtained from the greedy merge process \cite{sennrich2016neural}.

To generate different segmentations for one word, subword regularization methods draw a segmentation from the segmentation distribution probabilistically:
\begin{equation}
    \mathbf{x} \sim P_{seg}(\mathbf{x} | \mathbf{s};p).
\label{equ:subword-random}
\end{equation}
For example, \citet{kudo2018subword} makes use of a unigram language model to sample segmentations on, and \citet{provilkov2020bpe} randomly interrupts the BPE merging process to generate multiple segmentations.

\section{Deterministic Reversible Data Augmentation}

Previous data augmentation and subword regularization approaches take irreversible operation (like discrete token substitution) and probabilistic segmentation sampling, which may introduce semantic loss or inappropriate subwords, thus affecting the semantic consistency. Our objective is to ensure the semantic consistency between original and augmented data when generating diverse data.

We propose DRDA to generate augmentation data without introducing semantic perturbations. DRDA augments original data with multi-granularity segmentations, and pulls representations of one sentence closer with multi-view learning. Furthermore, we propose a dynamic selection technique to automatically choose an appropriate granularity in inference.

\begin{figure}[!htb]
    \centering
    \includegraphics[width=0.95\linewidth]{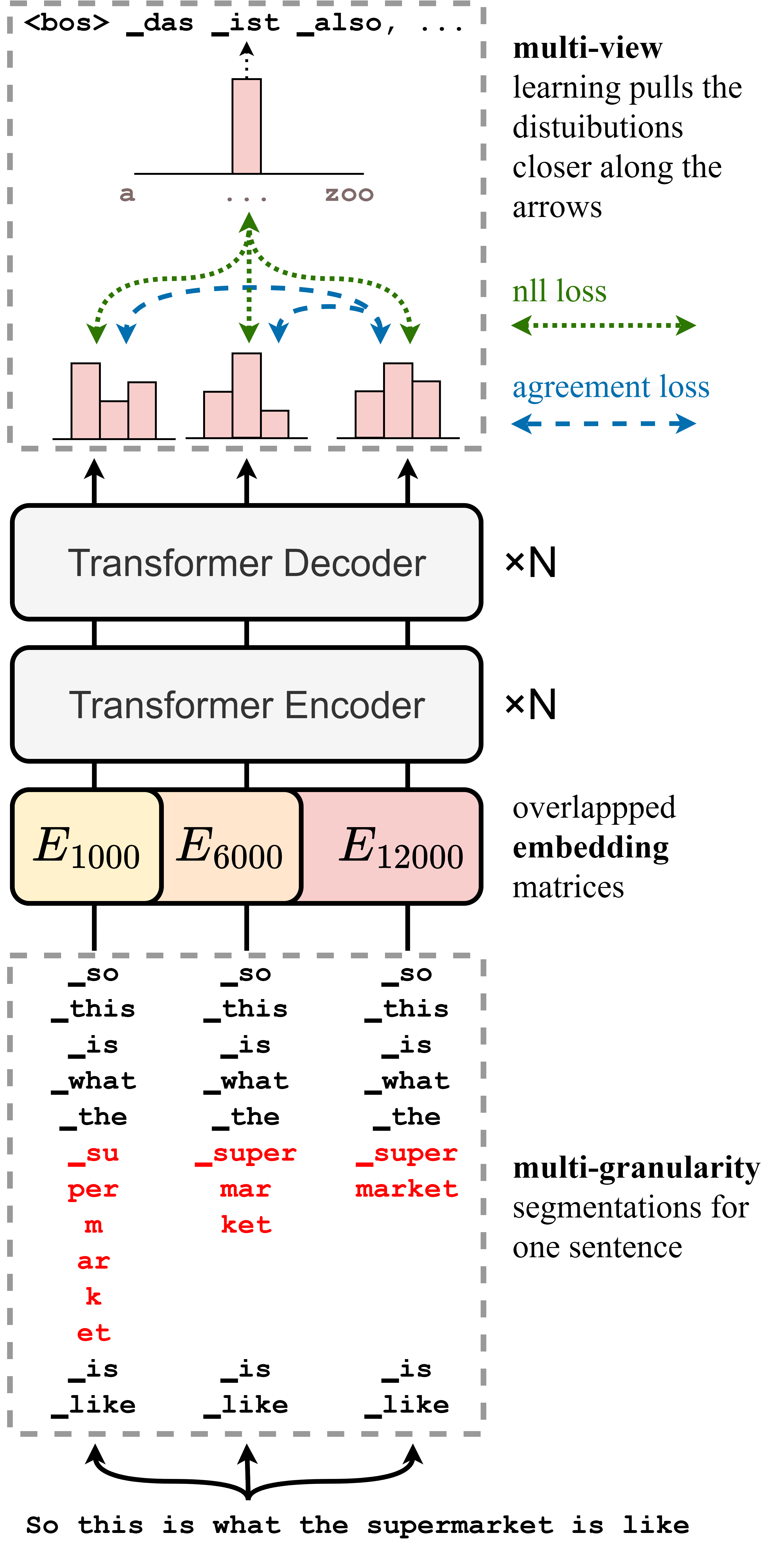}
    \caption{Illustration of the overall framework of DRDA. A source sentence is segmented into different granularities, and every generated token sequence will go through the model, obtaining a hypothesis distribution respectively. The agreement loss (blue segmented lines) will be computed between hypothesis distributions, and the negative likelihood loss (green dotted lines) will be computed between each distribution and the target.}
    \label{fig:model}
\end{figure}

\subsection{Multi-Granularity Segmentations}
\label{subsec:DRDA}

DRDA constructs symbolically diverse and semantically consistent augmentation data with multi-granularity segmentations. The point is that multi-granularity subword segmentation is a reversible process that completely retains semantic information, and is a deterministic process that always chooses the most probable and appropriate subword segmentation policy.

Formally, given a prime vocabulary size $p$ and a set of augmented vocabulary sizes $\{q_i\}_{i=1}^k$, for a source-target translation pair sample $(\mathbf{s}, \mathbf{t})$, a prime source sequence $\mathbf{x}^{pri}$, a target sequence $\mathbf{y}$ and a set of augmented source sequences $\{\mathbf{x}^{aug_i}\}_{i=1}^k$ can be generated:

\begin{equation}
    \mathbf{x}^{pri} = \mathop{\arg\max}\limits_{\mathbf{x} \in V_p} P(\mathbf{x} | \mathbf{s}),
\end{equation}
\begin{equation}
    \mathbf{x}^{aug_i} = \mathop{\arg\max}\limits_{\mathbf{x} \in V_{q_i}} P(\mathbf{x} | \mathbf{s}),
\end{equation}
\begin{equation}
    \mathbf{y} = \mathop{\arg\max}\limits_{\mathbf{y'} \in V_p} P(\mathbf{y'} | \mathbf{t}).
\end{equation}

Figure \ref{fig:model} depicts the model architecture and training loss on a English$\rightarrow$Germany sample. Given $p=12000$, $q_{1}=1000$, and $q_{2}=6000$, an English sentence is segmented with different vocabularies, generating three token sequences with different granularities.

Note that according to the greedy property of BPE, a short vocabulary is a prefix of a long vocabulary, as long as they are obtained from the same corpus. As a result, introducing different granularities with BPE will not lead to a larger vocabulary, thus avoiding an increase in parameter size. An example is shown in Figure \ref{fig:model}, where three embedding matrices $E_{12000}$, $E_{6000}$ and $E_{1000}$ are overlapped, and a smaller embedding is a prefix of a larger embedding.

\subsection{Multi-view Learning}

Moreover, to make the translation model learn from different segmentation granularities, we utilize the multi-view learning loss function \cite{wang2021multi, kambhatla2022cipherdaug} and pull different representations closer together:
\begin{equation}
    \begin{aligned}
        \mathcal{L} \ \ &= \ \ \underbrace{ \mathcal{L}_{NLL}(P(\mathbf{y} \vert \mathbf{x}^{pri}; \theta))}_{\rm{prime \ source \ loss}}
        \\
        &+ \ \ \underbrace{\frac{1}{k} \sum_{i=1}^k \mathcal{L}_{NLL}(P(\mathbf{y} \vert \mathbf{x}^{aug_i}; \theta))}_{\rm{augmented \ source \ loss}}
        \\
        &+ \ \ \underbrace{\frac{\alpha}{k} \sum_{i=1}^k \mathcal{L}_{dist}(P(\mathbf{y} \vert \mathbf{x}^{pri}; \theta), P(\mathbf{y} \vert \mathbf{x}^{aug_i}; \theta))}_{\rm{agreement \ loss}},
    \end{aligned}
\label{equ:loss}
\end{equation}
where $\mathcal{L}_{NLL}$ is the negative likelihood loss in machine translation, $\mathcal{L}_{dist}$ is the symmetric Kullback-Leibler divergence \cite{kambhatla2022cipherdaug}.

The first two terms of Equation \ref{equ:loss} (prime source loss and augmented source loss) compute the translation loss for source and augmented sentences respectively, and the third term (agreement loss) pulls the prediction distributions of different source inputs together.

As shown in Figure \ref{fig:model}, output probability distributions for all granularities are used to compute the loss, where the blue segmented lines refer to the agreement loss between different granularities, and green dotted lines refer to the negative likelihood loss between the prediction and the target.

\subsection{Dynamic Selection of Granularity in Inference}
\label{subsec:dynamic}

DRDA employs multiple segmentations in different granularities, so the selection of the granularity used in inference becomes a concern. To automatically choose a suitable vocabulary size when inferring, we also propose a simplified but granularity-focused version of $n$-best decoding \cite{kudo2018subword} to dynamically select the segmentation granularity in inferring step.

Given the set of all prime and augmented vocabulary sizes $\{p, q_1, q_2, \cdots, q_k\}$ and an input sentence $\mathbf{s}$, a series of $(\mathbf{x}, \mathbf{y})$ pairs can be generated, where each $(\mathbf{x}, \mathbf{y})$ pair represents a source-target token sequence pair in a certain granularity.

The estimated most probable segmentation and translation pair corresponds to the $(\mathbf{x}, \mathbf{y})$ pair that maximizes the following score:
\begin{equation}
    score(\mathbf{x}, \mathbf{y}) = \log P(\mathbf{y} \vert \mathbf{x}) / |\mathbf{y}|,
\end{equation}
where $|\mathbf{y}|$ is the length of $\mathbf{y}$.

\section{Experiments}

We evaluate DRDA with translation tasks in different language pairs and translation directions to show its universal property regardless of language features. We also conduct experiments on extremely low resources and noisy scenarios to show the robustness of DRDA.\footnote{Further setup details about dataset split, preprocessing, models, and evaluation are listed in Appendix \ref{sec:appendix1}.}

\subsection{Experimental Setup}
\label{subsec:experimental}

\begin{table}[htbp]
\centering
 \resizebox{\linewidth}{!}{
\begin{tabular}{cccc}
\toprule
 & \bf{WMT} & \bf{IWSLT} & \bf{TED} \\
 & \bf En $\rightarrow$ De & \bf En $\leftrightarrow$ (De, Fr, Zh, Es) & \bf En $\leftrightarrow$ Sk \\
 \midrule\midrule
 train & 4.5M & 160k, 236k, 235k, 183k & 61k \\
 valid & 3000 & 7283,  9487,  9428,  5593 & 2271 \\
 test  & 3003 & 6750,  1455,  1459,  1305 & 2445 \\
 \bottomrule
\end{tabular}
}
\caption{Overviews of datasets and corresponding sizes.}
\label{tab:datasets}
\end{table}

\begin{table*}[htbp]
    \centering
    \resizebox{\linewidth}{!}{
    \begin{tabular}{l|cccccccc|cc}
    \toprule
    \multirow{2}{*}{\bf{Model}} & \multicolumn{8}{c|}{\bf{IWSLT}} & \multicolumn{2}{c}{\bf{WMT}} \\
    & \bf{En}$\rightarrow$\bf{De} & \bf{De}$\rightarrow$\bf{En} & \bf{En}$\rightarrow$\bf{Fr} & \bf{Fr}$\rightarrow$\bf{En} & \bf{En}$\rightarrow$\bf{Zh} & \bf{Zh}$\rightarrow$\bf{En} & \bf{En}$\rightarrow$\bf{Es} & \bf{Es}$\rightarrow$\bf{En} & \bf{En}$\rightarrow$\bf{De} & \bf{De}$\rightarrow$\bf{En}\\
    \midrule\midrule
    Transformer      & 29.03 & 35.26 & 37.57 & 37.29 & 22.38 & 21.29 & 39.92 & 41.86 & 27.08 & 29.84 \\
    \midrule
    DRDA           & 30.84{\ddag} & 37.90{\ddag} & \bf{38.77}{\ddag} & \bf{38.55}{\dag} & \bf{23.36}{\dag} & 22.64{\dag} & 41.99{\ddag} & 43.90{\ddag} & 27.41{\dag} & 31.48{\ddag}\\
    \bf{DRDA dyn.} & \bf{30.92}{\ddag} & \bf{37.95}{\ddag} & 38.75{\dag} & 38.52{\dag} & 23.32{\dag} & \bf{22.90}{\dag} & \bf{42.07}{\ddag} & \bf{44.08}{\ddag} & \bf{27.45{\dag}} & \bf{31.59}{\ddag}\\
    \bottomrule
    \end{tabular}
    }
    \caption{BLEU on IWSLT and WMT. Statistical significance over Transformer is indicated by {\dag} ($p < 0.05$) and {\ddag} ($p < 0.001$). Significance is computed via bootstrapping \cite{koehn2004statistical} using \texttt{compare-mt} \cite{neubig2019compare}.}
    \label{tab:iwslt}
\end{table*}

\paragraph{Datasets and preprocessing}

Our experiments are conducted on different datasets, as detailed in Table \ref{tab:datasets}. We experiment on a low resource setting with IWSLT datasets, including IWSLT14 En$\leftrightarrow$De, En$\leftrightarrow$Es, and IWSLT17 En$\leftrightarrow$Zh, En$\leftrightarrow$Fr. We use larger WMT14 En$\leftrightarrow$De as a high-resource scenario dataset. The performance in extremely low resource scenarios is explored with the TED En$\leftrightarrow$Sk dataset. Following previous work \cite{vaswani2017attention}, we lowercase words in IWSLT En$\leftrightarrow$De, while keeping other datasets cased.\footnote{En, De, Fr, Zh, Es, Sk stand for English, German, French, Chinese, Spanish, and Slovak respectively.}

\paragraph{Models}

We build models on top of Transformer \cite{vaswani2017attention} with \texttt{Fairseq} toolkit \cite{ott2019fairseq}. We use a Base Transformer model \texttt{transformer\_wmt\_en\_de} for WMT, and \texttt{transformer\_iwslt\_de\_en} for others.

\paragraph{Hyperparameters in training and inferring}

We use \texttt{sentencepiece} \cite{kudo2018sentencepiece} to perform tokenization and BPE segmentation. The BPE encoding model is learned jointly on the source and target sides except for IWSLT En$\leftrightarrow$Zh. Unless otherwise stated, we use two vocabulary tables (on prime vocabulary and one augmented vocabulary), and their vocabulary sizes follow Table \ref{tab:vocab}. Detailed analysis of the vocabulary sizes and the number of augmented vocabularies will be shown in Section \ref{sec:ablations}. The weight of agreement loss $\alpha$ is set to 5 unless otherwise stated.

\begin{table}[htbp]
    \centering
    % \resizebox{0.8\linewidth}{!}{
    \begin{tabular}{cccc}
    \toprule
    & WMT & IWSLT & TED \\
    \midrule\midrule
    DRDA pri & 32k & 10k & 8k \\
    DRDA aug & 16k & 5k  & 4k \\
    others   & 32k & 10k & 8k \\
    \bottomrule
    \end{tabular}
    % }
    \caption{Prime and augmented vocabulary sizes used in DRDA, and vocabulary sizes used in other methods.}
    \label{tab:vocab}
    \end{table}

\paragraph{Evaluation}

We evaluate the performance of NMT systems using BLEU. To compare with previous work \cite{vaswani2017attention,kambhatla2022cipherdaug}, we apply multi-bleu with \texttt{multi\_bleu.perl} \footnote{mosesdecoder/scripts/generic/multi-bleu.perl} for IWSLT En$\leftrightarrow$De, WMT En$\rightarrow$De, and TED En$\leftrightarrow$Sk. For WMT En$\rightarrow$De dataset, we additionally apply compound splitting\footnote{tensorflow/tensor2tensor/utils/get\_ende\_bleu.sh}. All other datasets are evaluated with \texttt{SacreBLEU}\footnote{SacreBLEU signature: \texttt{nrefs:1|case:mixed| eff:no|tok:13a|smooth:exp|version:2.2.0}}.

\subsection{Main Result}

We present the results of DRDA on IWSLT and WMT translation tasks in Table \ref{tab:iwslt}. We can see that DRDA consistently outperforms the Transformer with a clear margin on all translation tasks. Moreover, models inferred with the dynamic granularity selection obtain a modest improvement in DRDA.

\begin{table}[htb]
\centering
% \resizebox{\linewidth}{!}{
\begin{tabular}{l|cc}
\toprule
\bf{Model} & \bf{En}$\rightarrow$\bf{De} & \bf{De}$\rightarrow$\bf{En} \\
\midrule\midrule
Transformer & 29.03  & 35.26 \\
\midrule
WordDrop    & 29.21 & 35.60 \\
SwitchOut   & 29.00 & 35.90 \\
RAML        & 29.70 & 35.99 \\
DataDiverse & 30.47 & 37.00 \\
BPE-Drop    & 30.16 & 36.54  \\
SubwordReg  & 29.46 & 36.14  \\
R-Drop      & 30.45 & 37.40  \\
MVR         & 30.44 & 37.47 \\
CipherDAaug & 30.65  & 37.60 \\
\midrule
DRDA            & 30.84{\ddag} & 37.90{\ddag} \\
\bf{DRDA dyn.} & \bf{30.92}{\ddag} & \bf{37.95}{\ddag}\\
\bottomrule
\end{tabular}
% }
\caption{BLEU scores on IWSLT En$\leftrightarrow$De. Results of previous data augmentation (the second to the fifth models) are cited from literature which we share the same configuration with, as detailed in Appendix \ref{sec:appendix1}.}
\label{tab:iwsltdeen}
\end{table}

Comparison between DRDA and other data augmentation and subword regularization methods on IWSLT are shown in Table \ref{tab:iwsltdeen}. We use a range of augmentation and regularization methods for comparison. The augmentation methods include WordDrop \cite{zhang2020token, sennrich2016edinburgh}, SwitchOut \cite{wang2018switchout}, RAML \cite{norouzi2016reward} and Data Diversification \cite{nguyen2020data}. The subword regularization methods include BPE-Drop \cite{provilkov2020bpe} and Subword Regularization \cite{kudo2018subword}. We also compare our method with others that adopt multi-view learning techniques, including R-Drop \cite{wu2021r}, MVR \cite{wang2021multi}, and CipherDAug \cite{kambhatla2022cipherdaug}. DRDA yields greater improvement compared to others.

\subsection{Extremely Low Resource Setting}

TED En$\leftrightarrow$Sk task is challenging because of its extremely low resources (only 61k training sentence pairs). Several techniques have been adopted to improve the performance in low-resource NMT tasks like this, including data augmentation, multilingual translation, and transfer learning \cite{ranathunga2021neural}. \citet{neubig2018rapid} firstly propose similar language regularization to mix low-resource language with a lexically related high-resource language, combining transfer learning and multilingual translation. Several works continue to extend SRL and achieve high translation quality \cite{xia2019generalized, ko2021adapting, wang2018multilingual}.

\begin{table}[htbp]
\centering
\begin{tabular}{l|cc}
\toprule
\bf{Model} & \bf{En}$\rightarrow$\bf{Sk} & \bf{Sk}$\rightarrow$\bf{En}\\
\midrule\midrule
Transformer & 20.82  & 28.97 \\
\midrule
LRL+HRL     & -      & 32.07\\
CipherDAaug & 24.61 & 32.62\\
\midrule
DRDA        & 24.48  & 33.25\\
\bf{DRDA dyn.} & \bf{24.67} & \bf{33.34}\\
\bottomrule
\end{tabular}
\caption{BLEU scores on TED En$\leftrightarrow$Sk. LRL+HRL method combines the original low-resource language pair with a high-resource related language Czech.}
\label{tab:ted}
\end{table}

On this task, DRDA yields stronger improvements over baseline Transformer than other techniques with no requirement for external high resource languages, as shown in Table \ref{tab:ted}.

\subsection{Robustness to Perturbations}
\label{subsec:robust}

We validate the robustness of DRDA on two noisy datasets. The first one is IWSLT De$\rightarrow$En test set with synthetic perturbations. The perturbations are synthesized by traversing every character excluding space and punctuation in source sentences, and applying one of the operations with probability $0.01$: (1) remove the character, (2) add a random character following the character, and (3) substitute the character with a random one. The second dataset is himl test set\footnote{https://www.himl.eu/test-sets}, which contains health information and scientific summaries and differs considerably from the IWSLT training set. Cross-domain datasets have different subword distributions, and the difference can be viewed as a natural noise. The results of the noisy test sets are shown in Table \ref{tab:noisy}.

\begin{table}[htbp]
\centering
% \resizebox{\linewidth}{!}{
\begin{tabular}{l|c|cc}
\toprule
& original & synthetic & himl\\
\midrule\midrule
Transformer & 35.26 & 32.19 & 26.11 \\
\midrule
R-Drop      & 37.40 & 34.34 & 28.15 \\
BPE-Drop    & 36.54 & \bf{35.00} & 27.92 \\
SubwordReg  & 36.14 & 34.55 & 27.63 \\
\midrule
DRDA      & 37.90 & 34.94 & \bf{28.80} \\
DRDA dyn. & \bf{37.95} & 34.98 & 28.78 \\
\bottomrule
\end{tabular}
% }
\caption{BLEU scores on original and noisy IWSLT De$\rightarrow$En test set, and himl test set. Models are trained on the IWSLT De$\rightarrow$En training set.}
\label{tab:noisy}
\end{table}

Along with these results, consistent improvement over Transformer and R-Drop is obtained by DRDA on both synthetically noisy and cross-domain datasets. DRDA significantly outperforms subword sampling methods (BPE-Drop and subword regularization) on natural noise datasets, but only obtains similar results with synthetic noise. We will discuss the reason in Section \ref{subsec:consistency}.

\section{Analysis}

In this section, we conduct analysis experiments to answer the following research questions (RQs) respectively:

\begin{itemize}%[noitemsep]
    \item RQ1 (ablation studies): How do the applied techniques and components affect model performance?
    \item RQ2: Does our approach really keeps semantic consistency between original and augmented data?
    \item RQ3: How does multi-granularity segmentation improve subword representations?
    \item RQ4: Why does multi-view learning help improve NMT models?
\end{itemize}

\subsection{RQ1: Ablations}
\label{sec:ablations}

\begin{figure}[htbp]
    \centering
    \includegraphics[width=1.0\linewidth]{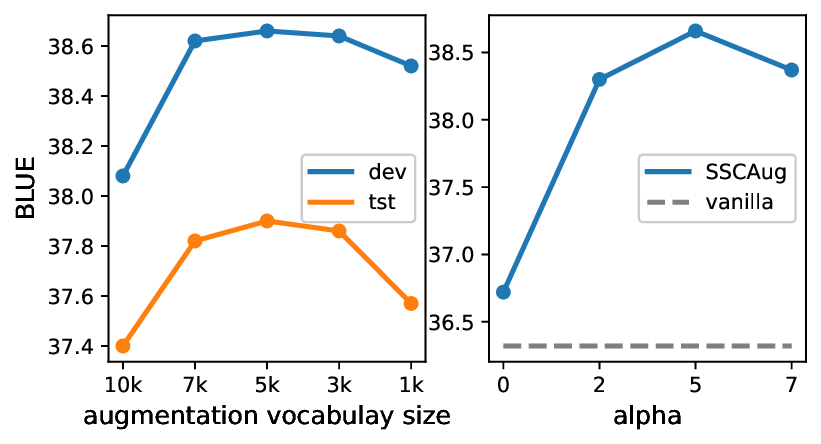}
    \caption{Ablations on IWSLT De$\rightarrow$En over augmented vocabulary size (left) and agreement loss weight (right).}
    \label{fig:ablation}
\end{figure}

\paragraph{Choice of vocabulary sizes}

Here, we investigate the effects of pre-defined vocabulary sizes. As is mentioned in Section \ref{subsec:experimental}, we adopt one prime vocabulary and one augmented vocabulary. To find the optimal vocabulary sizes, we test \{10k, 7k, 5k, 3k, 1k\} for augmented vocabulary size when the prime size is 10k. Figure \ref{fig:ablation} verifies that, when the augmented vocabulary size is around 5k, the NMT model obtains the highest BLEU. The intuition is a huge difference in prime and augmented vocabulary sizes may corrupt the subword semantics, while a tiny difference may reduce the symbolic difference. A general recommendation in choosing vocabulary sizes is to use a proven suitable size for the prime vocabulary and set the augmented size to half the size of the prime vocabulary.

\paragraph{Weight of agreement loss}

As is shown in Figure \ref{fig:ablation}, we find that agreement loss weight $\alpha$ significantly affects the performance of our method. Models obtain the highest BLEU score when $\alpha=5$, and increasing or decreasing $\alpha$ causes a score drop up to 2 BLEU on the valid set. The model without agreement loss (i.e., $\alpha = 0$) still outperforms vanilla Transformer, validating the important role multi-granularity segmentation plays in DRDA.

\paragraph{Number of augmented vocabularies}

\begin{table}[htb!]
    \centering
    \begin{tabular}{ccc|cc}
    \toprule
    \bf{\{1k\}} & \bf{\{6k\}} & \bf{\{9k\}} & $\mathbf{\mu}$ & $\mathbf{\sigma}$ \\
    37.90 & 38.17 & 37.95 & 38.01 & 0.12 \\
    \midrule
    \bf{\{1k,6k\}} & \bf{\{1k,9k\}} & \bf{\{6k,9k\}} & $\mathbf{\mu}$ & $\mathbf{\sigma}$ \\
    38.21 & 38.20 & 38.16 & 38.19 & 0.02 \\
    \bottomrule
    \end{tabular}
    \caption{BLEU scores on IWSLT De$\rightarrow$En valid set when a 12k prime vocabulary is combined with different augmented vocabulary sets. $\mu$ and $\sigma$ refer to the mean and standard deviation of BLEU scores when combined with one (top) or two (bottom) augmented vocabularies.}
    \label{tab:ntables}
\end{table}

Table \ref{tab:ntables} shows the effects of adding an extra augmented vocabulary with a prime vocabulary size of 12k on the valid set. When combined with two augmented vocabularies, the BLEU scores have a smaller deviation than combined with one. We can summarize that adding extra augmented vocabularies helps get a steady, comparable, and maybe slightly better result in the cost of an increase in training time.

\subsection{RQ2: Semantic Consistency}
\label{subsec:consistency}

\paragraph{Theoretical discussion}

In this section, we discuss what is semantic consistency, and give a theoretical analysis about why DRDA is more semantically consistent.

It is clear that previous data augmentation methods that adopt irreversible operations result in semantic loss, which will inevitably do damage to the consistency between original and augmented data. DRDA is superior to these methods in terms of preservation of the original meanings, because it is based on reversible segmentation to generate diversity.

However, it is more challenging to prove that subword regularization methods \cite{kudo2018subword,provilkov2020bpe}, which are also based on reversible segmentation, lead to greater inconsistency than DRDA. To show the superiority of DRDA in consistency over subword regularization, we review the difference of the two in sampling segmentation:
\begin{equation}
    \mathbf{x}_{DRDA}^i = \mathop{\arg\max}\limits_{\mathbf{x}} P_{seg}(\mathbf{x} | \mathbf{s};p_i),
\end{equation}
\begin{equation}
    \mathbf{x}_{SR} \sim P_{seg}(\mathbf{x} | \mathbf{s};p),
\end{equation}
where $\mathbf{x}_{DRDA}^i$ is a representation in certain granularity of source sentence $s$ in DRDA, $\mathbf{x}_{SR}$ is the representation in subword regularization, $p_i$ and $p$ are vocabulary sizes.

$\mathop{\arg\max}\limits_{\mathbf{x}} P_{seg}(\mathbf{x} | \mathbf{s};p)$ can be interpreted as the difficulty of segmenting $s$ with a certain vocabulary size $p$. We can assume that the difficulty of segmenting a sentence is an inherent property of sentences, independent of vocabulary sizes:
\begin{equation}
    \mathop{\arg\max}\limits_{\mathbf{x}} P_{seg}(\mathbf{x} | \mathbf{s}) = \mathop{\arg\max}\limits_{\mathbf{x}} P_{seg}(\mathbf{x} | \mathbf{s};p),
\end{equation}
where $p \in \mathbb{N}$ is any pre-defined vocabulary size.

Then, because of the deterministic argmax operation in DRDA and the random sampling operation in subword regularization, the following inequality holds:
\begin{equation}
    P_{seg}(\mathbf{x}_{DRDA}|\mathbf{s}) \ge P_{seg}(\mathbf{x}_{SR} | \mathbf{s}).
    \label{equ:theo}
\end{equation}

Equation \ref{equ:theo} validates that our approaches generates more appropriate segmentations of a same sentence that other subword regularization methods. As a result, although both DRDA and subword regularization are reversible, DRDA is semantically more consistent because of the segmentation appropriateness.

\paragraph{Empirical analyses}

To give an empirical insight of the semantical consistency, we analyze the nearest neighbors of subwords of different models (shown in Table \ref{tab:neighbour}). We can find that vanilla Transformer and DRDA both exhibit semantics-based neighbors, where the embeddings of synonyms are similar. However, embeddings obtained in BPE-Drop tend to have high similarity with those they share a common sequence. Although this tendency can effectively alleviate vulnerability to misspelling, which explains the superiority subword regularization shows in synthetic noisy data in Section \ref{subsec:robust}, it may introduce semantic error as well (treat "\texttt{\_go}" and "\texttt{\_god}" as synonyms for "\texttt{\_good}" in Table \ref{tab:neighbour} for example), causing inaccuracy in machine translation.

\begin{table}[htbp]
    \centering
    \begin{tabular}{ccc}
    \toprule
    \multicolumn{3}{c}{\texttt{\_good}} \\
    Transformer & DRDA & BPE-Drop \\
    \midrule
    \texttt{\_great} & \texttt{\_great} & \texttt{\_great} \\
    \texttt{\_better} & \texttt{\_big} & \texttt{\_go} \\
    \texttt{\_nice} & \texttt{\_nice} & \texttt{\_bad} \\
    \texttt{\_bad} & \texttt{\_bad} & \texttt{\_god} \\
    \texttt{\_useful}& \texttt{\_significant} & \texttt{\_nice} \\
    \bottomrule
    \end{tabular}
    \caption{Top 5 nearest neighbors of subwords "\texttt{\_good}" on IWSLT De$\rightarrow$En.}
    \label{tab:neighbour}
\end{table}

The observation above indicates that DRDA introduces little semantic noise to augmentation data, and exhibits better semantic consistency.

\subsection{RQ3: Effects on Subword Frequency}
\label{subsec:freq}

Here, we show that the mechanism of multi-granularity segmentation can be attributed to the increase in frequency of infrequent tokens.

\begin{figure}[!htb]
    \centering
    \includegraphics[width=1\linewidth]{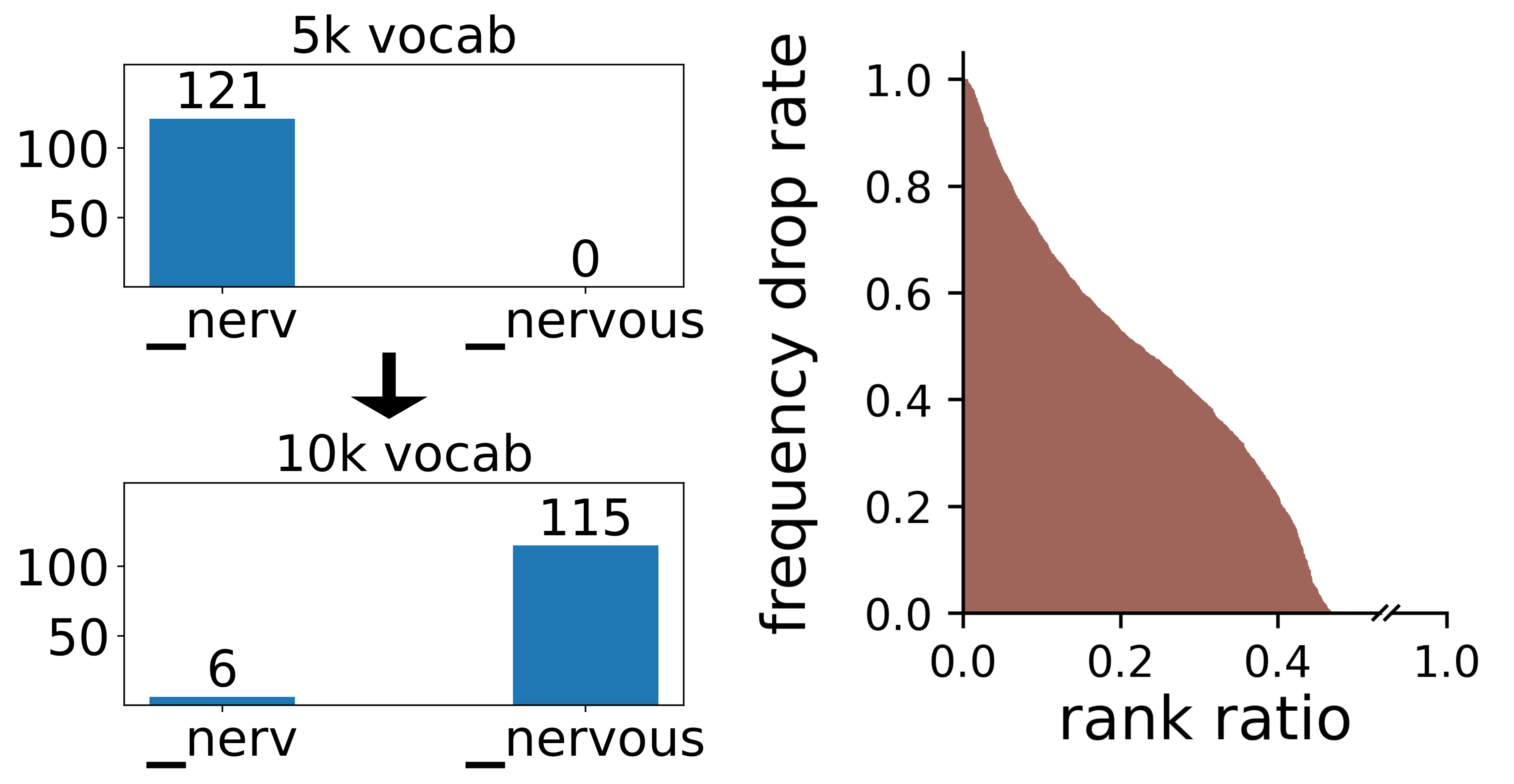}
    \caption{Most occurrences of "\texttt{\_nerv}" are absorbed by "\texttt{\_nervous}" when the vocabulary grows (left). The frequency drop rate of "\texttt{\_nerv}" is $(121-6)/121 = 0.95$. The right figure shows all frequency drop rates on IWSLT En$\rightarrow$De sorted in descending order.}
    \label{fig:contexts}
\end{figure}

NMT models with larger vocabulary sizes have larger atomic translation units, i.e., more coarse-grained subwords, so that they can better memorize one-to-many or many-to-one mappings and resolve translation ambiguity \cite{koehn2009statistical}. However, fine-grained subwords may suffer from a frequency drop when the vocabulary size grows. Figure \ref{fig:contexts} shows that most occurrences of "\texttt{\_nerv}" are absorbed by "\texttt{\_nervous}" when the vocabulary grows, making it more difficult for the NMT model to obtain a precise representation of other inflection forms like "\texttt{\_nervy}", "\texttt{\_nervier}" and "\texttt{\_nervine}". More generally, the frequency drop is common on IWSLT En$\rightarrow$De (results on more datasets are shown in Appendix \ref{sec:appendix2}), where about 50\% of subwords appeared in 5k vocabulary suffer from a frequency drop when the vocabulary grows to 10k, as Figure \ref{fig:contexts} shows.

In DRDA, by taking both small and large vocabulary sizes simultaneously, infrequent tokens occur more frequently so that subwords like "\texttt{\_nerv}" can be trained in adequate contexts as well.

\subsection{RQ4: Multi-view Techniques and Subword Semantic Composition}
\label{subsec:ssc}

Multi-view learning pulls representations in different granularities together. To investigate the effects of multi-view techniques, we propose a task to find out how the coarse-grained and fine-grained representations of the same word are drawn closer.

\begin{figure}[!htb]
    \centering
    \includegraphics[width=0.8\linewidth]{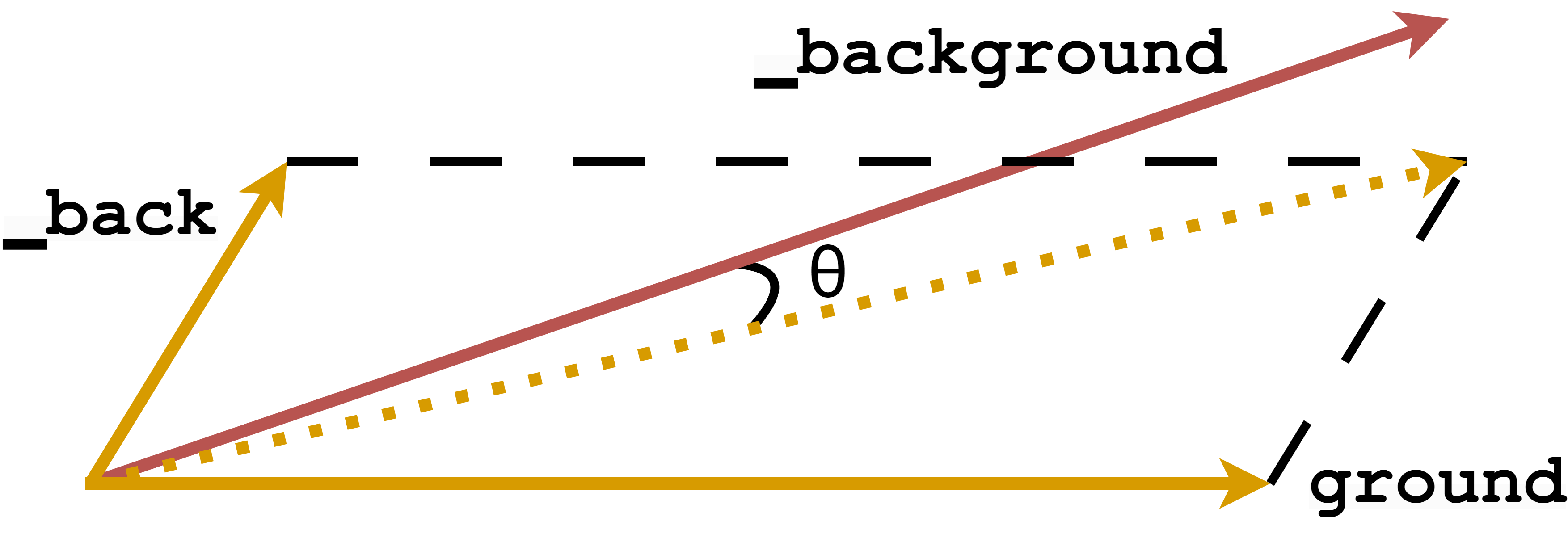}
    \caption{The similarity between the fine- and coarse-grained representations is computed by $\cos \theta$.}
    \label{fig:ssc}
\end{figure}

The process is illustrated with an example in Figure \ref{fig:ssc}, and the formal definition of the task is shown in Appendix \ref{sec:ssc-task}. We take a coarse-grained subword ("\texttt{\_background}") and its corresponding fine-grained subword sequence ("\texttt{\_back}", "\texttt{ground}"), then compute the cosine similarity between the former embedding and the sum of the latter embeddings. The similarity indicates the extent to which the fine-grained and coarse-grained representations are brought closer together.

We enumerate all the coarse-grained and fine-grained representation pairs, and average all their cosine similarity scores. The results are shown in Table \ref{tab:semantics}. As expected, DRDA with proper agreement loss ($\alpha=5$) obtains a higher average similarity than other data augmentation approaches.

\begin{table}[htbp]
    \centering
    \resizebox{\linewidth}{!}{
    \begin{tabular}{l|ccc|c}
    \toprule
    \multirow{2}{*}{\bf{Model}} & $\texttt{\_back},$ & $\texttt{\_plat},$ & $\texttt{\_feed},$ & \multirow{2}{*}{{\bf \texttt{avg}}} \\
    & $\texttt{ground}$ & $\texttt{form}$ & $\texttt{back}$ & \\
    \midrule\midrule
    Transformer & 0.22 & 0.28 & 0.31 & 0.24 \\
    R-Drop      & 0.41 & 0.36 & 0.39 & 0.35 \\
    BPE-Drop    & 0.54 & 0.62 & 0.66 & 0.46 \\
    \midrule
    DRDA $\alpha=0$ & 0.48 & 0.50 & 0.66 & 0.50 \\
    DRDA $\alpha=2$ & 0.71 & 0.67 & 0.77 & 0.67 \\
    DRDA $\alpha=5$ & 0.78 & \bf{0.79} & \bf{0.85} & \bf{0.77} \\
    DRDA $\alpha=7$ & \bf{0.81} & 0.74 & 0.82 & 0.75 \\
    \bottomrule
    \end{tabular}
    }
    \caption{Similarities between coarse- and fine-grained representations for the same word (e.g., "\texttt{\_background}" vs. "\texttt{\_back}"+"\texttt{ground}"). {\bf \texttt{avg}} refers to the average similarities of all words on IWSLT En$\rightarrow$De.}
    \label{tab:semantics}
\end{table}

Computing the similarities between representations in multiple granularities is a subword level composition (SSC) tasks \cite{mitchell2008vector, mitchell2009language, turney2014semantic}. We can conclude that multi-view techniques help DRDA models improve the SSC understanding, thus obtaining better robustness to perturbations \cite{provilkov2020bpe}.

\section{Conclusion}

In this paper, we identify the semantic inconsistency caused by irreversible operations or probabilistic segmentations, and propose a deterministic reversible data augmentation consisting of multi-granularity segmentation and multi-view learning to ensure the consistency when generating diverse data. Experiments demonstrate the superiority of our proposed DRDA over previous data augmentation and subword regularization in terms of translation accuracy and robustness. We also offer a combination of empirical and theoretical verification of semantic consistency, and insightful analyses about multi-granularity and multi-view techniques.

\section*{Limitations}

\paragraph{High resource scenarios}
As other data augmentation techniques, our proposed DRDA appears to be less effective in high-resource scenarios (up to 1.75 BLEU gain in WMT, and 2.69 in IWSLT) than in low resource scenarios (up to 4.37 BLEU gain in TED). The analysis in Section \ref{subsec:freq} offers one explanation to this phenomenon that, the frequency drop becomes less sharp when the data size grows, thus resulting in lower effectiveness of data augmentation. Considering this phenomenon, a better application approach of data augmentation on high-resource scenarios can be designed, by locating the rare subwords of a specific domain in a model trained on large general corpus and continuing training with the augmentation data. We leave this investigation as a direction for future research.

\paragraph{Application scope}
As a foundation process in NLP, segmentation is applied in various tasks, including language modeling, named entity recognition, and numerous others. Additionally, vision tasks like image translation can also benefit from segmentation \cite{tian-etal-2023-image}. As a result, segmentation based data augmentation techniques including DRDA can be applied to a wide range of tasks. One limitation of this study is its exclusive application of DRDA to machine translation, which restricts the ability to validate and compare its effectiveness across other tasks.

\section*{Acknowledgements}
We'd like to thank all the anonymous reviewers for their diligent efforts in helping us improve this work. This work is supported by the National Natural Science Foundation of China (Grant No. U21B2009) and Beijing Institute of Technology Science and Technology Innovation Plan (Grant No. 23CX13027).

% % Entries for the entire Anthology, followed by custom entries
\bibliography{anthology,custom}
\bibliographystyle{acl_natbib}

\appendix

\section{Implementation Details}
\label{sec:appendix1}

\subsection{Datasets and Preprocessing}

We perform minimum preprocessing and cleaning steps to raw data.

\begin{itemize}[noitemsep]

    \item For IWSLT En$\leftrightarrow$De, following previous works, data is obtained with \texttt{fairseq} scripts\footnote{fairseq/example/translation//prepare\_iwslt14.sh}, which performs \texttt{clean-corpus-n} \footnote{mosesdecoder/scripts/training/clean-corpus-n.perl} with ${\rm{ratio}}=1.5$, ${\rm{min}}=1$ and ${\rm{max}}=175$.

    \item For other IWSLT datasets, we extract titles, descriptions and main texts for training, and main texts only for validating and testing. There is no extra cleanup operation performed. IWSLT14 En$\leftrightarrow$Es dataset concatenates \texttt{dev2010}, \texttt{tst2010}, \texttt{tst2011} and \texttt{tst2012} as development set, uses \texttt{tst2015} as test set. IWSLT17 En$\leftrightarrow$Fr and En$\leftrightarrow$Zh datasets concatenate \texttt{dev2010}, \texttt{tst2010}, \texttt{tst2011}, \texttt{tst2012}, \texttt{tst2013}, \texttt{tst2014} and \texttt{tst2015} as development set, use \texttt{tst2017} as test set.

    \item We use \texttt{t2t-datagen} \footnote{tensorflow/tensor2tensor/bin/t2t-datagen} script to generate WMT data, and performs \texttt{clean-corpus-n} with ${\rm{min}}=1$ and ${\rm{max}}=80$, removing about 1\% training sentence pairs. Following previous works, we validate on \texttt{newstest2013} and test on \texttt{newstest2014}.

    \item The TED datasets are obtained using scripts from the official repository \cite{qi2018and}. We additionally remove the encoder language token "\texttt{\_\_sk\_\_}" to accommodate bilingual NMT.

\end{itemize}

\subsection{Models and Hyperparameters}

Smaller datasets are trained with model \texttt{transformer\_iwslt\_de\_en}, and WMT dataset is trained with model \texttt{transformer\_wmt\_en\_de}. The corresponding config is shown in Table \ref{tab:config}.

\begin{table}[htb!]
    \centering
    % \resizebox{\linewidth}{!}{
    \begin{tabular}{c|cc}
    \toprule
    & small & base \\
    \midrule\midrule
    encoder layer & 6 & 6  \\
    decoder layer & 6 & 6  \\
    attention head  & 4 & 8  \\
    embedding size  & 512 & 512  \\
    feed-forward size  & 1024 & 2048  \\
    learning rate & $6e-4$ & $5e-4$ \\
    lr schedule & inverse sqrt & inverse sqrt \\
    optimizer & adam & adam \\
    adam betas & (0.9, 0.98) & (0.9, 0.98) \\
    clip norm & 0.1 & - \\
    warm updates & 8000 & 4000 \\
    network dropout & 0.3 & 0.1 \\
    attention dropout & 0.1 & 0.1 \\
    weight decay & $1e-4$ & 0 \\
    label smoothing & 0.1 & 0.1 \\
    words per batch & about 17k & about 380k \\
    beam & 5 & 4 \\
    length penalty & 1.0 & 0.6 \\
    \bottomrule
    \end{tabular}
    % }
    \caption{Model configuration of \texttt{transformer\_iwslt\_de\_en} (small) and \texttt{transformer\_wmt\_en\_de} (base).}
    \label{tab:config}
\end{table}

Note that DRDA, R-Drop, CipherDAug and some other approaches may double the input texts, but we constrain the tokens number forwarded to the model in a batch according to the "words per batch" hyperparameter, which means the numbers of sentences in a batch of these approaches are rough halved.

\subsection{Computational Cost}

Total training duration and GPU used in DRDA experiments are listed in Table \ref{tab:cost}.

\begin{table}[htbp]
    \centering
    \resizebox{\linewidth}{!}{
    \begin{tabular}{c|ccc}
    \toprule
    & WMT & IWSLT & TED \\
    \midrule\midrule
    \multirow{2}{*}{GPU}  & $2 \times$ & $4 \times$ & $2 \times$ \\
        & RTX 3090 & TITAN Xp & TITAN X (Pascal) \\
    \midrule
    time & 12 days & 2 days & 10 hours \\
    steps & 100k & 200k & 35k \\
    \bottomrule
    \end{tabular}
    }
    \caption{Computational cost of WMT, IWSLT and TED experiments.}
    \label{tab:cost}
\end{table}

\subsection{Baseline Implementation}

We reimplement those models with high relevance with our method, including vanilla Transformer, BPE-Drop, subword regularization, R-Drop, MVR and CipherDAaug. These models except for Transformer use either segmentation-related techniques or multi-view techniques. Important details of our implementation are listed below:

\begin{itemize}[noitemsep]
    \item BPE-Drop and subword regularization are implemented using \texttt{sentencepiece}. In encoding, we set $\alpha=0.1$ and $\alpha=0.2$ for BPE-Drop and subword regularization respectively, and ${\rm{nbest\_size}}=-1$ for both. Results of subword regularization are obtained without $n$-best decoding \cite{kudo2018subword}.
    \item We use the task and loss module from the official open-source repository \footnote{https://github.com/dropreg/R-Drop} to implement R-Drop. Weight $\alpha$ is set to 5.
    \item In MVR implementation, we adopt the same subword regularization hyper-parameters to BPE-Drop, and the agreement loss weight is set to be $5$.
    \item CipherDAaug models are reimplemented on top of the official open-source code\footnote{https://github.com/protonish/cipherdaug-nmt}. Following their instructions, we adopt 2 keys, and set agreement loss weight $\beta = 5$.
\end{itemize}

For the traditional data augmentation methods ( WordDrop, SwitchOut, RAML, DataDiverse) with which DRDA shares a relatively low similarity, results are cited from \citet{kambhatla2022cipherdaug}. We share exactly the same model architecture and hyperparameters with \citet{kambhatla2022cipherdaug}, and we successfully reimplemented their main model with similar results, so we find it reliable to cite from. 

We report the performance of LRL+HRL from the corresponding literature \cite{xia2019generalized}.

\section{Process of Subword Semantic Composition Task}
\label{sec:ssc-task}

Let $a \circ b$ be a compound token concatenated by $a$ and $b$, with their corresponding embedding ${\bf{e}}_{a \circ b}$, ${\bf{e}}_{a}$ and ${\bf{e}}_{b}$, the SSC understanding ability is scored by the similarity between ${\bf{e}}_{a \circ b}$ and ${\bf{e}}_{a}+{\bf{e}}_{b}$:
\begin{equation}
   {\rm{SIM}}({\bf{e}}_{a \circ b}, {\bf{e}}_{a}+{\bf{e}}_{b}) = \frac{{\bf{e}}_{a \circ b} \cdot ({\bf{e}}_{a}+{\bf{e}}_{b})}{ \Vert {\bf{e}}_{a \circ b} \Vert \cdot \Vert {\bf{e}}_{a}+{\bf{e}}_{b} \Vert }.
\end{equation}

To numerically evaluate the superiority of a model in understanding SSC, we average semantic composition similarities of all subwords except characters and special tokens (such as \texttt{<unk>}):
\begin{equation}
\overline{{\rm{SIM}}} = \frac{1}{|\widetilde{V}|} \sum\limits_{a, b, a \circ b \in V}{\rm{SIM}}({\bf{e}}_{a \circ b}, {\bf{e}}_{a}+{\bf{e}}_{b}),
\end{equation}
where $\widetilde{V}$ is a set of all subwords except characters and special tokens, and $V$ is a set of all subwords.

It should be noted that the models listed in Section \ref{subsec:ssc} share the same $V$, so that comparing the scores completely makes sense.

\section{More Studies}
\label{sec:appendix2}

\subsection{Subword Nearest Neighbors}

\begin{table}[htb!]
    \centering
    \resizebox{\linewidth}{!}{
    \begin{tabular}{cc|cc}
    \toprule
    \multicolumn{2}{c}{\texttt{\_good}} &\multicolumn{2}{c}{\texttt{ood}} \\
    DRDA & BPE-Drop & DRDA & BPE-Drop \\
    \midrule
    \texttt{\_great} & \texttt{\_great} & \texttt{oods} & \texttt{oo} \\
    \texttt{\_big} & \texttt{\_go} & \texttt{\_penguins} & \texttt{oods} \\
    \texttt{\_nice} & \texttt{\_bad} & \texttt{ago} & \texttt{ook} \\
    \texttt{\_bad} & \texttt{\_god} & \texttt{\_birthday} & \texttt{od} \\
    \texttt{\_significant} & \texttt{\_nice} & \texttt{wow} & \texttt{\_food} \\
    \texttt{\_better} & \texttt{\_useful} & \texttt{ghter} & \texttt{\_good} \\
    \texttt{\_huge} & \texttt{ood} & \texttt{\_entrop} & \texttt{wood} \\
    \texttt{\_happy} & \texttt{\_goods} & \texttt{anced} & \texttt{ool} \\
    \texttt{\_useful} & \texttt{\_better} & \texttt{gal} & \texttt{\_blood} \\
    \texttt{\_healty} & \texttt{\_big} & \texttt{\_astero} & \texttt{idung} \\
    \midrule\midrule
    \multicolumn{2}{c}{\texttt{\_photograph}} &\multicolumn{2}{c}{\texttt{\_photographs}} \\
    DRDA & BPE-Drop & DRDA & BPE-Drop \\
    \midrule
    \texttt{\_photo} & \texttt{\_phot} & \texttt{\_pictures} & \texttt{\_photos} \\
    \texttt{\_photos} & \texttt{\_photographs} & \texttt{\_photos} & \texttt{\_pictures} \\
    \texttt{\_photographs} & \texttt{\_photo} & \texttt{\_images} & \texttt{\_photograph} \\
    \texttt{\_picture} & \texttt{\_fotograf} & \texttt{\_movies} & \texttt{\_phot} \\
    \texttt{\_pattern} & \texttt{\_picture} & \texttt{\_structures} & \texttt{ć} \\
    \texttt{\_digit} & \texttt{\_ph} & \texttt{\_chemicals} & \texttt{\_images} \\
    \texttt{\_penguins} & \texttt{\_graph} & \texttt{\_statistics} & \texttt{ť} \\
    \texttt{\_tremend} & \texttt{ograph} & \texttt{\_maps} & \texttt{ē} \\
    \texttt{\_pictures} & \texttt{ť} & \texttt{\_scene} & \texttt{\_} \\
    \texttt{\_doctors} & \texttt{\_photos} & \texttt{\_spectrum} & \texttt{©} \\
    \bottomrule
    \end{tabular}
    }
    \caption{Top 10 nearest neighbors of example subwords.}
    \label{tab:neighbours}
\end{table}

Following \citet{provilkov2020bpe}, we study the closest neighbors of word embedding learned in BPE-Drop and DRDA. Several examples are shown in Table \ref{tab:neighbours}.

We can find that in the morphology of words, BPE-Drop tends to bring two subwords sharing a common sequence together ("\texttt{\_good}" and "\texttt{ood}" for example), while DRDA has no such behavior. On one hand, the tendency to pull similarly spelled words closer can effectively help NMT model overcome the perturbation of misspelling, as shown in previous experiments. On the other, it can introduce unreasonable noise as well, since similarly spelled subwords are not necessarily semantically related words ("\texttt{\_good}" and "\texttt{\_go}" for example).

\subsection{Effects of Granularity Selection}

Our experiments have shown that the dynamic selection of segmentation granularity yields a modest improvement in BLUE scores. Here, we investigate the mechanism and potential of this method.

\begin{table}[htbp]
    \centering
     \resizebox{\linewidth}{!}{
    \begin{tabular}{c|cccc}
    \toprule
     & prime & augmented & dynamic & oracle \\
    \midrule\midrule
    \bf{En} $\rightarrow$ \bf{De} & 30.84 & 30.85 & 30.92 & 31.36 \\
    \bf{De} $\rightarrow$ \bf{En} & 37.90 & 37.88 & 37.95 & 38.46 \\
    \bf{En} $\rightarrow$ \bf{Fr} & 38.77 & 38.57 & 38.75 & 40.15 \\
    \bf{Fr} $\rightarrow$ \bf{En} & 38.55 & 38.44 & 38.52 & 39.59 \\
    \bf{En} $\rightarrow$ \bf{Zh} & 23.36 & 23.32 & 23.32 & 23.37 \\
    \bf{Zh} $\rightarrow$ \bf{En} & 22.64 & 22.84 & 22.90 & 23.62 \\
    \bf{En} $\rightarrow$ \bf{Es} & 41.99 & 42.07 & 42.07 & 42.64 \\
    \bf{Es} $\rightarrow$ \bf{En} & 43.90 & 43.97 & 44.08 & 45.30 \\
    \bottomrule
    \end{tabular}
    }
    \caption{BLEU scores on IWSLT tasks.}
    \label{tab:dynamic}
\end{table}

We define an oracle granularity selection model, whose translation result corresponds to the one with the highest sentence-level BLEU score among the results generated by source sequences with different granularities. The results of models with 5k augmented size and 10k prime size on IWSLT translation tasks are shown in Table \ref{tab:dynamic}.

It can be summarized from the results that the selection of input granularities has considerable potential (up to 1.7 BLEU) in improving the translation, and our approach obtains an improvement of up to 0.24 BLEU. In the future, a better re-ranking approach can be adopted to build a selection model closer to the oracle model.

\subsection{Frequency Drops}
\label{subsec:more-drops}

More examples of frequency drop on WMT, IWSLT and TED are shown in Figure \ref{fig:freqdrops}. Among these results, all datasets suffer from a similar frequency drop regardless of their language directions and sizes. The vocabulary size grows from 16k to 32k for WMT, from 5k to 10k for IWSLT and from 4k to 8k for TED.

\section{Discussion on Potential Application in Language Modeling}

As a foundation process in NLP, segmentation is applied in various tasks. As a result, segmentation based data augmentation techniques including DRDA can also be applied to a wide range of tasks including language modeling. Considering the significant upsurge in the development of large language models (LLMs) \cite{brown2020language, achiam2023gpt} and their relevance to machine translation \cite{hendy2023good}, we discuss the potential application of DRDA in language modeling.

In our preliminary view, DRDA may be beneficial in language modeling on training or fine-tuning steps. Specifically, consider the scenario where a language model is trained using the IWSLT dataset. In such instances, the model will encounter the issue described in Section \ref{subsec:freq}, where the limited contextual information can hinder its ability to accurately capture the diverse inflectional forms of a word due to the frequency drop phenomenon. Analogous to its application in machine translation, DRDA could potentially mitigate this frequency drop, thus improving the word representation learning.

\begin{figure}[!htb]
    \centering
    \includegraphics[width=0.95\linewidth]{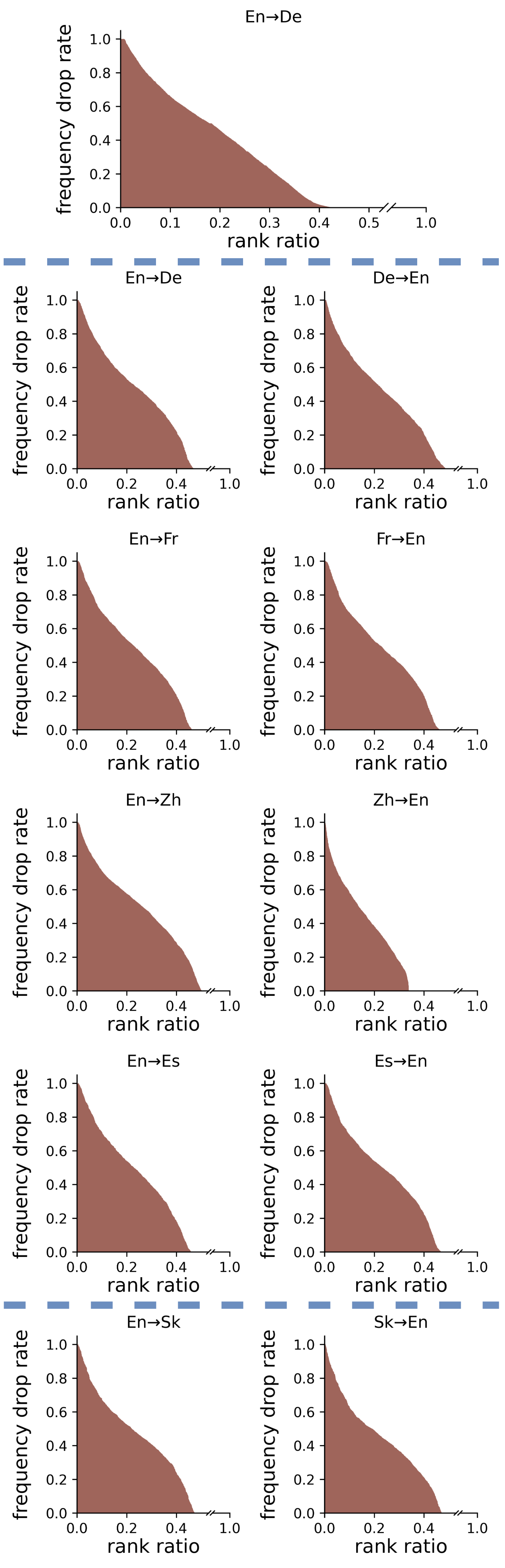}
    \caption{Subwords frequency drop rates on WMT (top), IWSLT (middle), and TED (bottom).}
    \label{fig:freqdrops}
\end{figure}

However, in term of LLMs trained with huge amount of data, the frequency drop phenomenon is observed to be less significant. This is evident from  the fact that the larger WMT dataset exhibit a more moderate frequency drop than IWSLT, as shown in Figure \ref{fig:freqdrops}. Considering this, we believe DRDA is particularly beneficial in a specific downstream scope where there exists rare and complex terminologies, but the training corpus is not sufficient for the model to understood those words. For example, in the context of biology, the term \texttt{glycosaminoglycan} may occur infrrequently. Nevertheless, by leveraging the decomposition of this term into \texttt{glyco}, \texttt{amino}, and \texttt{glycan}, which are more frequently encountered in the corpus, the model can potentially develop a deeper understanding of the term.

\end{document}